\pdfoutput=1

\documentclass[11pt]{article}

\usepackage[preprint]{acl}

\usepackage{times}
\usepackage{latexsym}
\usepackage{hyperref}
\usepackage{url}
\makeatletter
\g@addto@macro{\UrlBreaks}{\do\/\do\-}
\makeatother
\usepackage[T1]{fontenc}

\usepackage[utf8]{inputenc}

\usepackage{microtype}

\usepackage{inconsolata}

\usepackage{graphicx}
\usepackage{microtype}
\usepackage{amsmath}
\usepackage{amssymb}
\usepackage{algpseudocode}
\usepackage{booktabs}
\usepackage{algorithm}
\usepackage{xcolor}
\usepackage{xpatch}
\makeatletter
\xpatchcmd{\algorithmic}
  {\ALG@tlm\z@}{\leftmargin\z@\ALG@tlm\z@}
  {}{}
\makeatother

\title{Memory-Efficient Backpropagation for Fine-Tuning LLMs on Resource-Constrained Mobile Devices}

\author{Congzheng Song \\
  Apple \\
  \texttt{csong4@apple.com} \\\And
  Xinyu Tang \\
  Apple \\
  \texttt{xinyu\_tang3@apple.com} \\}

\begin{document}
\maketitle

\begin{abstract}
Fine-tuning large language models (LLMs) with backpropagation\textemdash even for a subset of parameters such as LoRA~\citep{hu2022lora}\textemdash can be much more memory-consuming than inference and is often deemed impractical for resource-constrained mobile devices.
Alternative methods, such as zeroth-order optimization (ZO), can greatly reduce the memory footprint but come at the cost of significantly slower model convergence (10× to 100× more steps than backpropagation).
We propose a memory-efficient implementation of backpropagation (MeBP) on mobile devices that provides better trade-off between memory usage and compute time, while converging faster and achieving better performance than the ZO baseline.
We verify the effectiveness of MeBP on an iPhone 15 Pro Max and show that various LLMs, ranging from 0.5B to 4B parameters, can be fine-tuned using less than 1GB of memory.
We release an example of the MeBP implementation at \url{https://github.com/apple/ml-mebp}.
\end{abstract}

\section{Introduction}
Large language models (LLMs) have been successfully integrated into mobile devices to run inference on users' private data locally~\citep{gunter2024apple,geminiteam2025geminifamilyhighlycapable}.
For applications such as personalization or federated learning~\citep{mcmahan2017communication}, it is also desirable to fine-tune models on local private data on-device to further improve utility~\citep{kairouz2021advances}.
However, fine-tuning LLMs with backpropagation on mobile devices remains extremely challenging due to the significantly higher memory footprint compared to inference.
These on-device training processes typically run in the background, which further limits memory usage due to operating system constraints~\cite{apple,android}.
In addition, total training compute time must be short to prevent the OS from interrupting or rescheduling the training process.

Existing works on memory-efficient on-device fine-tuning of LLMs have focused on approximating gradients with zeroth-order optimization (ZO)~\citep{Spall1992MultivariateSA}, such as MeZO~\citep{malladi2023finetuning}, where the memory footprint is similar to standard vanilla inference, as no backpropagation is required.
While ZO methods help reduce memory usage in theory, ZO often suffers from slower and poorer convergence, leading to longer compute times and degraded model performance (Section~\ref{sec:eval}).
Even with ZO, existing implementations require multiple gigabytes of memory to train a small LLM (e.g., OPT-1.3B~\citep{zhang2022opt}), which is impractical for any production deployment~\cite{peng2024pocketllm}.

In this work, we present a memory-efficient implementation of backpropagation (MeBP) for fine-tuning LLMs on mobile devices.
The implementation is based on gradient checkpointing~\citep{chen2016training}, with various optimizations including lazy weight loading and decompression, as well as memory-mapped activation checkpoints.
Our implementation ensures that no extra intermediate activations or uncompressed base model weights are kept in memory\textemdash they are only loaded when computation is needed.
The total training memory footprint is thus reduced to that of backpropagation on a single checkpoint, which is feasible within the RAM constraints of mobile devices.

We implement MeBP in iOS using Swift and evaluate its performance on an iPhone 15 Pro Max.
We focus on a language modeling task and compare MeBP with MeZO on a set of LLMs suitable for deployment on mobile devices, including Gemma3~\citep{team2025gemma} and Qwen2.5~\citep{qwen2024qwen25technicalreport}.
We demonstrate that MeBP converges faster and better than MeZO in terms of both the number of optimization steps and total compute wall-clock time.
In addition, MeBP incurs only a slightly higher memory footprint than MeZO, making it more practical for on-device training.

\section{Related Works}

\paragraph{Memory efficient training.} Training machine learning models incurs memory costs from model parameters, gradients, optimizer states, and intermediate values like activations. Each of these components offers opportunities for optimization to reduce memory usage during training.
Prior works have proposed base model quantization~\citep{dettmers2023qlora} and CPU offloading~\citep{zero} to reduce the memory cost of model parameters.
To reduce the memory cost of computing gradients, parameter-efficient fine-tuning (PEFT) methods such as LoRA~\citep{hu2022lora} reduce trainable parameters to less than $1\%$ of the total model parameters. These PEFT methods significantly lower gradient-related memory usage and achieve competitive performance compared to full model training for fine-tuning tasks.
In-place weight updates with gradients during backpropagation—instead of updating model parameters after completing all backpropagation steps—can also reduce gradient memory cost~\citep{lv2024lomo}.
Prior works~\citep{dettmers2022bit,zhao2024galore} have also studied how to reduce the GPU memory cost for optimizer states such as AdamW~\citep{kingma2014adam} under full-model training.

Reducing the memory cost of gradients, optimizer states, and intermediate activations can help narrow the memory usage gap between model training and vanilla model inference.
Gradient checkpointing~\citep{chen2016training} significantly reduces the memory cost of intermediate activations by trading off memory usage for increased computation time through recomputation during backpropagation.
\citet{malladi2023finetuning} proposed a memory-efficient version of zeroth-order optimization, MeZO, which estimates gradients via seeded random perturbations and therefore incurs only negligible additional memory cost compared to vanilla inference.
However, zeroth-order fine-tuning typically requires significantly more ($10\times$ to $100\times$) optimization steps than first-order methods.
Several follow-up works~\citep{qin2024federated,zhao2025secondorder,dang2025fzoo} have been proposed to improve the convergence rate of MeZO.

\paragraph{On-device training.}
On-device training enables machine learning models to adapt to on-device data while preserving data privacy.
\citet{lin2022ondevice} fine-tuned a small convolutional neural network on tiny IoT devices with limited SRAM (e.g., 256KB) using quantization, PEFT methods, and system-algorithm co-design.
For language models with billions of parameters, PocketLLM~\citep{peng2024pocketllm} uses MeZO for on-device fine-tuning of LLMs, but it still incurs significant memory costs (6.5GB for OPT-1.3B~\citep{zhang2022opt}), which is impractical for mobile devices.

\section{Memory-Efficient Backpropagation}
\begin{algorithm*}[t!]
\footnotesize
\caption{Memory-Efficient Backpropagation}
\label{alg:MeBP}
\begin{algorithmic}
\State \textbf{Inputs:} input data $x$, number of checkpoints $n$, forward checkpoint subgraphs [\texttt{forward}$_i$], backward checkpoint subgraphs [\texttt{backward}$_i$], LoRA trainable weights [\texttt{lora\_weights}$_i$] for each checkpoints, compressed base model weights for each checkpoints [\texttt{compressed\_base\_weights}$_i$]
\State
\Procedure{\textnormal{\texttt{InitializeModel}}}{}
\State Memory map (mmap) all weights in [\texttt{compressed\_base\_weights}$_i$]
\EndProcedure
\State
\Procedure{\textnormal{\texttt{LazyLoadAndDecompressWeights}}}{$i$}
\State Load mmaped \texttt{compressed\_base\_weights}$_i$ for checkpoint index $i$
\State \Return $\texttt{decompress}(\texttt{compressed\_base\_model\_weights}_i)$
\EndProcedure
\State
\Procedure{\textnormal{\texttt{Backpropagation}}}{$x$}
\State Initialize \texttt{ckpts\_storage} $\gets \{x\}$
\State Load current LoRA trainable weights [\texttt{lora\_weights}$_i$]
\For{each checkpoint index $i\in[1, \dots, n]$}  \Comment{\textit{\color{gray}Forward pass to store all checkpoints}}
\State Load \texttt{base\_weights}$_i\gets$\texttt{LazyLoadAndDecompressWeights}($i$)
\State Load mmaped \texttt{ckpts}$_{i-1}$ from \texttt{ckpts\_storage}
\State Compute \texttt{ckpts}$_i\gets\texttt{forward}_i(\texttt{lora\_weights}_i,\texttt{base\_weights}_i, \texttt{ckpts}_{i-1})$
\State Mmap \texttt{ckpts}$_i$ and add to \texttt{ckpts\_storage}
\EndFor
\State Initialize \texttt{lora\_grads}$\gets\emptyset$, \texttt{ckpts\_grads}$_{n+1}\gets$\texttt{nil}
\For{each checkpoint index $i\in[n, \dots, 1]$}  \Comment{\textit{\color{gray}Backward pass in reverse order to compute gradients}}
\State Load \texttt{base\_weights}$_i\gets$\texttt{LazyLoadAndDecompressWeights}($i$)
\State Load mmaped \texttt{ckpts}$_{i}$ from \texttt{ckpts\_storage}
\State Compute (\texttt{lora\_grads}$_i$, \texttt{ckpts\_grads}$_i$) $\gets\texttt{backward}_i(\texttt{lora\_weights}_i, \texttt{base\_weights}_i, \texttt{ckpts}_{i}, \texttt{ckpts\_grads}_{i+1})$
\State Remove \texttt{ckpts}$_i$ from \texttt{ckpts\_storage}
\State Update \texttt{lora\_grads}$\gets\texttt{lora\_grads}\cup \{\texttt{lora\_grads}_i\}$
\EndFor
\State \Return \texttt{lora\_grads}
\EndProcedure
\end{algorithmic}
\end{algorithm*}

We focus on fine-tuning LLMs with LoRA~\citep{hu2022lora} in this paper.
Therefore, the main memory bottlenecks lie in the model parameters and intermediate activations.
Our goal is to keep the memory usage of fine-tuning within a reasonable range for a modern mobile device (e.g., less than 1GB, as suggested by PocketLLM~\cite{peng2024pocketllm}).

There are three steps for fine-tuning LLMs with memory-efficient backpropagation (MeBP) on-device: 
1) compressing the model base weights (frozen parameters) to reduce disk space; 
2) compiling the training graph with backpropagation and gradient checkpointing for memory optimization; and 
3) implementing a memory-efficient runtime for executing the compiled training graph.
We describe each step in detail below.

\paragraph{Base model weights compression.}
It is common practice to compress base model weights to reduce disk space usage when deploying LLMs on-device.
In our implementation, we use 4-bit symmetric mode INT4 quantization on non-LoRA parameters including the embeddings.
We leave the investigation of more aggressive compression methods, such as 2-bit quantization-aware training~\citep{liu2025paretoq}, to future work.

\paragraph{Gradient checkpointing compilation.}
To implement gradient checkpointing in MeBP, we begin by splitting the LLM into blocks where the memory of  backpropagation on a single block (e.g. a transformer layer) is within the device memory constraints. 
For each block $F$ producing activations to be checkpointed, we generate the backward graph by applying automatic differentiation on the output of $F$.
For example, let $y = F_i(x, w)$ be the forward graph for block $F_i$, we perform automatic differentiation on the scalar $s$:
\begin{align*}
s &= \sum (\frac{\partial E}{\partial y} \odot y),
\\
\frac{\partial s}{\partial x} &= \frac{\partial E}{\partial y}\cdot\frac{\partial y}{\partial x} = \frac{\partial E}{\partial x}.
\end{align*}
where $E$ denotes the final loss to be optimized.
We can then produce a backward graph $(\frac{\partial E}{\partial x}, \frac{\partial E}{\partial w}) = B_i(x, \frac{\partial E}{\partial y}, w)$ where $\odot$ denotes Hardmard product and $\frac{\partial E}{\partial y}$ is outputted by the backward graph $B_{i+1}$. 
In other words, the inputs to the backward graphs are the checkpointed activations, gradients for the previous checkpoint and the corresponding trainable weights, and the outputs are the gradients of those inputs.
The forward and backward graphs for all blocks are then serialized into a device runtime compatible format, e.g. Model Intermediate Language (MIL) representation\footnote{\url{https://apple.github.io/coremltools/docs-guides/source/model-intermediate-language.html}} or MLX exported function\footnote{\url{https://ml-explore.github.io/mlx/build/html/python/export.html}}.
During runtime, the serialized graphs will be deserialized and compiled for computation.

\paragraph{Runtime implementation.}
\begin{figure*}[t]
    \centering
    \includegraphics[width=0.49\linewidth]{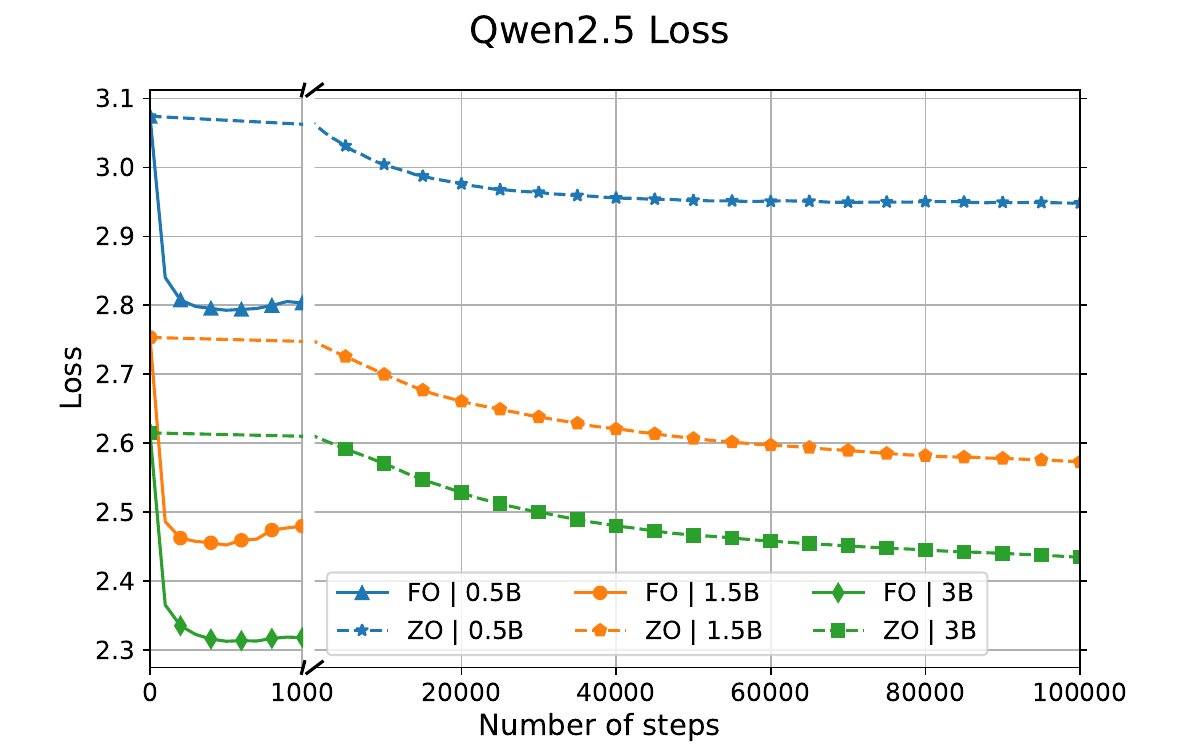}
    \includegraphics[width=0.49\linewidth]{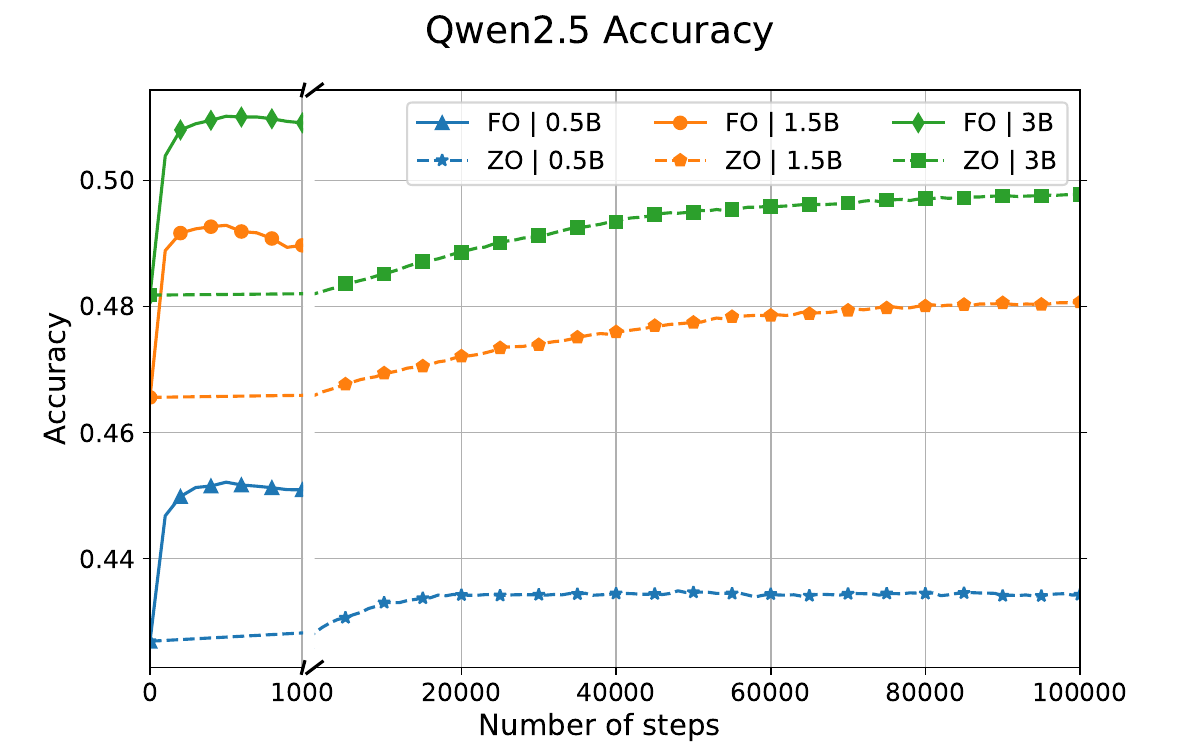}
    \includegraphics[width=0.49\linewidth]{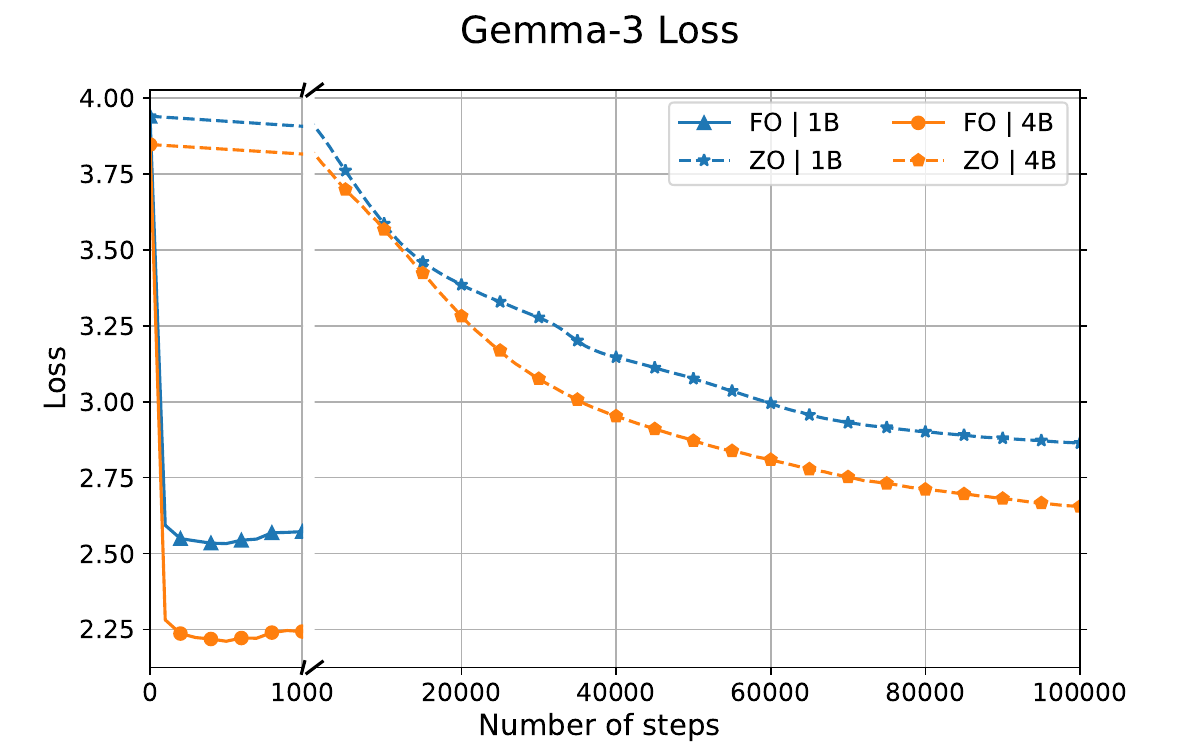}
    \includegraphics[width=0.49\linewidth]{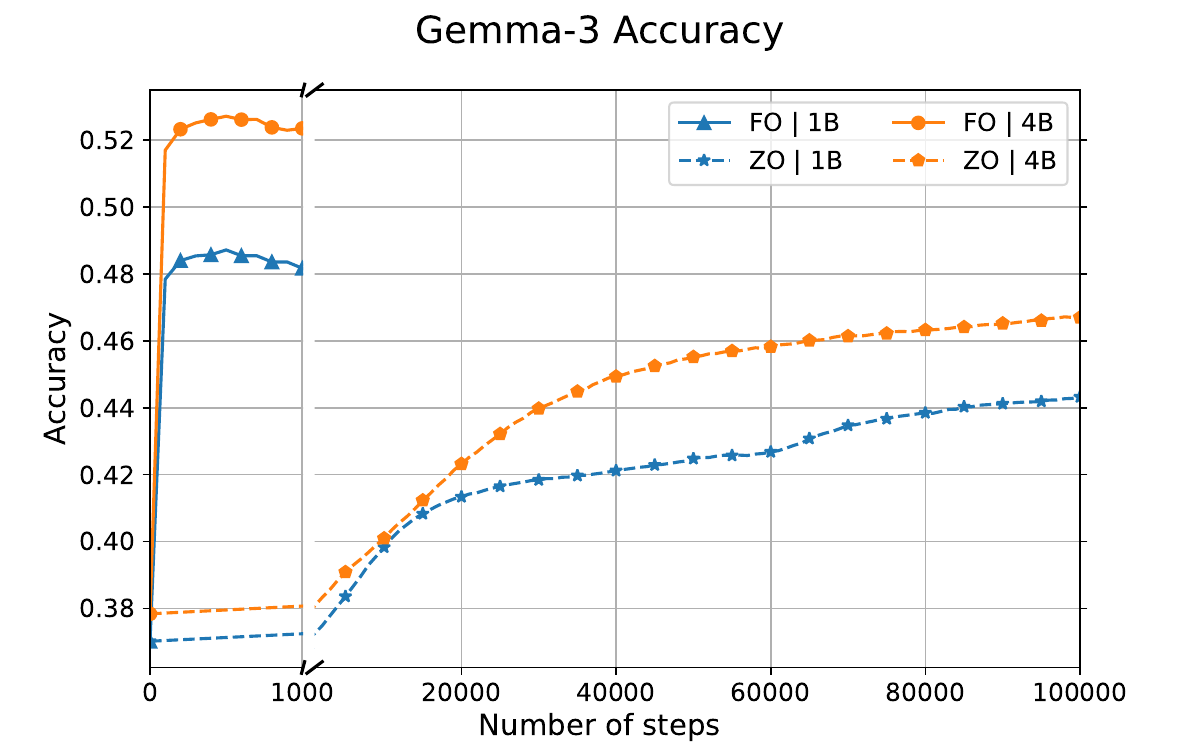}
    \caption{Convergence of Qwen2.5 (0.5B, 1.5B and 3B) and Gemma-3 (1B and 4B) fine-tuned with ZO and FO.}
    \label{fig:qwen2.5}
\end{figure*}

Algorithm~\ref{alg:MeBP} outlines the runtime implementation of MeBP.
The model is first initialized using the \texttt{InitializeModel} function, after which the \texttt{Backpropagation} function is invoked for each data point in the training loop.
During \texttt{InitializeModel}, the compressed base model weights are memory-mapped.
To minimize memory footprint, the base model weights are not decompressed before the training loop begins.
Instead, they are lazily decompressed and loaded on demand whenever required for computation.
Note that for device runtime frameworks supporting computation with quantized weights\footnote{\url{https://ml-explore.github.io/mlx/build/html/python/_autosummary/mlx.core.quantized_matmul.html}}, the decompression step can be skipped and only the compressed weights will be loaded on demand.

In the \texttt{Backpropagation} function, the forward compiled subgraphs are executed to store all necessary checkpoints, followed by the backward compiled subgraphs, which are executed in reverse order to compute the gradients using the stored checkpoints.
The checkpoints are memory-mapped during the forward pass rather than kept in memory.
Before each forward and backward pass, only the necessary base model weights are decompressed and loaded.
As a result, the total memory usage is limited to the size of the required base model weights plus the peak memory usage for operations in each subgraph which is significantly less than the full size of the base model weights.
The function describes gradient computation for a single data point. For batched inputs, gradient accumulation can be used to compute the gradient without increasing the memory footprint.

In MeBP, only a copy of the LoRA weights and their gradients is kept in memory for the optimizer.
For LLMs ranging from 0.5B to 4B parameters, the size of the LoRA weights is typically in the range of dozens of megabytes, which is reasonable to store in memory.
Optimizer states, such as momentum, can be memory-mapped and lazily loaded in a manner similar to the base model weights.

\begin{table*}[t]
\centering
\begin{tabular}{l|r|rr|rr}
\toprule
  &  & \multicolumn{2}{c|}{Time (s)} & \multicolumn{2}{c}{Memory (MB)}  \\
Model &  \# of trainable params & MeZO & MeBP & MeZO & MeBP \\
\midrule
Qwen2.5 0.5B & 4.39M & 2.68 & 3.85 & 318.93 & 320.17 \\
Qwen2.5 1.5B & 9.23M & 5.47 & 9.09 & 451.57 & 460.24 \\
Qwen2.5 3B & 14.97M & 10.28 & 17.96 & 554.10 & 661.78 \\
\midrule
Gemma3 1B & 6.52M &  4.88 & 9.48 & 563.64 & 569.00 \\
Gemma3 4B & 14.90M & 16.86 & 28.58 & 961.54 & 1029.49 \\
\bottomrule
\end{tabular}
\caption{Per-gradient-step compute time and peak memory of MeZO and MeBP.}
\label{tab:performance}
\end{table*}

\section{Experiments}
\label{sec:eval}
We consider MeZO as the baseline for demonstrating the performance of MeBP, as it is the only known optimization approach applied to LLM fine-tuning on mobile devices~\cite{peng2024pocketllm}.
We evaluate the utility of MeZO and MeBP through simulation on the server side and compare their performance on a mobile device, as detailed in the sections below.

\subsection{Utility Comparison}
\label{sec:zoo}
\paragraph{Setup.}
We compare the utility of first-order (FO) optimization (i.e., gradients via backpropagation) and zeroth-order (ZO) optimization by conducting experiments on the WikiText-2 dataset~\citep{merity2017pointer} for language modeling tasks using Gemma-3 and Qwen-2.5.
We focus on models with no more than 4B parameters, as mobile devices have constrained computing resources.
Our evaluation metrics are loss and next-token accuracy on the evaluation set.
Each sample has a sequence length of 256. We use a subset of the original WikiText-2 training set, consisting of 2,048 samples.
LoRA fine-tuning is applied in all experiments, with a rank of 16.
The total number of training steps is 1,000 for FO experiments and 100,000 for ZO experiments.
We use the AdamW optimizer for all experiments.
These experiments are run on the server side as a simulation to compare utility only.

\paragraph{Results.}
As shown in Figure~\ref{fig:qwen2.5}, while the loss and next-token accuracy for ZO exhibit a convergence trend, ZO converges significantly more slowly than FO.
The FO method improves both metrics substantially within the first 100 steps, whereas ZO shows only a slight improvement after 1,000 steps.
Even after 100,000 steps (i.e. $100\times$ more optimization steps than FO), ZO still yields higher test loss and lower test accuracy than FO for the same model.

Several methods have been proposed to improve the convergence rate of ZO methods~\citep{qin2024federated,zhao2025secondorder,dang2025fzoo}.
We also ran experiments using these improved ZO methods on Qwen2.5-0.5B and present the results in Figure~\ref{fig:zoo_improve} in Appendix~\ref{appendix:zo}.
While these methods achieve faster convergence than vanilla ZO, the loss and next-word token accuracy still remain worse than those of FO fine-tuned models.
Moreover, these methods typically require more computation time per iteration due to additional forward passes needed for more accurate gradient estimation.

The utility results demonstrate that backpropagation converges significantly faster than ZO methods for fine-tuning LLMs on language modeling tasks, on a per-step basis. This makes it more suitable for mobile deployment in terms of compute time, provided that each FO optimization step is implemented efficiently.

\subsection{Performance Comparison}
\paragraph{Setup.}
We implement MeBP in iOS using Swift and evaluate its performance on an iPhone 15 Pro Max, which has 8GB of RAM.
For the MeZO baseline implementation, the forward graph is split into multiple subgraphs, and lazy decompression is applied to reduce the total memory usage of the base model weights.
Each MeZO optimization step involves two forward passes.
We set the batch size to 1 and the sequence length to 256.
We checkpoint the model at every transformer layer, the final linear layer, and the cross-entropy loss layer.
Memory usage is recorded using the iOS native function \texttt{task\_vm\_info\_data\_t}, which provides the peak memory footprint of the running process via \texttt{phys\_footprint}.
We repeat the training process 10 times and report the average runtime and peak memory usage.

\paragraph{Results.}
Table~\ref{tab:performance} summarizes the performance results.
Overall, MeBP incurs 43\% to 94\% more computation time \textbf{per gradient step} compared to MeZO.
However, given that MeZO requires more than $10\times$ to $100\times$  the number of steps compared to first-order optimization as shown in the previous utility comparison, MeBP converges much faster in terms of wall-clock time.
MeBP uses up to 20\% more memory than MeZO in the worst case, while the total memory usage for training is approximately $10\times$ smaller than in previous mobile device implementations~\cite{peng2024pocketllm}.
All tested LLMs can be efficiently fine-tuned within 1GB of memory, making them suitable for background training on a mobile phone.

\begin{table}[t]
\centering
\begin{tabular}{l|rr}
\toprule
Model & Forward & Backward \\
\midrule
Qwen2.5 0.5B & 34.91\% & 15.80\% \\
Qwen2.5 1.5B & 32.77\% & 17.86\% \\
Qwen2.5 3B & 36.15\% & 21.15\% \\
Gemma3 1B & 32.37\% & 13.27\% \\
Gemma3 4B & 42.87\% & 24.18\% \\
\bottomrule
\end{tabular}
\caption{Ratio of decompression time during each forward and backward pass.}
\label{tab:decomp}
\end{table}

\begin{table}[t]
\centering
\begin{tabular}{l|rr|rr}
\toprule
Sequence & \multicolumn{2}{c|}{Time (s)} & \multicolumn{2}{c}{Memory (MB)}  \\
length & MeZO & MeBP & MeZO & MeBP \\
\midrule
128 & 4.81 & 6.92 & 367.49 & 405.14 \\
256 & 5.47 & 9.09 & 451.57 & 460.24 \\
512 & 9.61 & 17.14 & 617.82 & 624.62 \\
1024 & 18.18 & 34.40 & 986.00 & 994.09 \\
\bottomrule
\end{tabular}
\caption{Impact of sequence length.}
\label{tab:seq}
\end{table}

\begin{figure*}[ht]
    \centering
    \includegraphics[width=\linewidth]{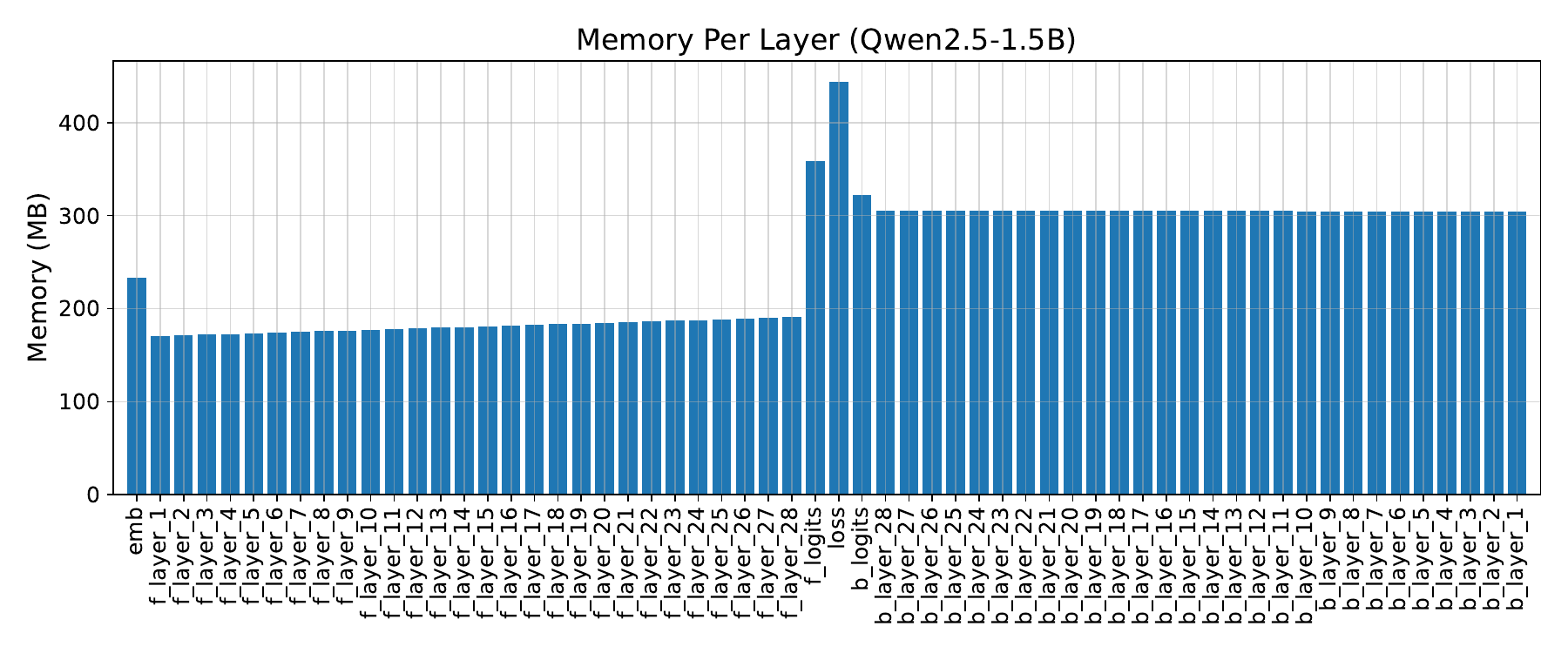}
    \includegraphics[width=\linewidth]{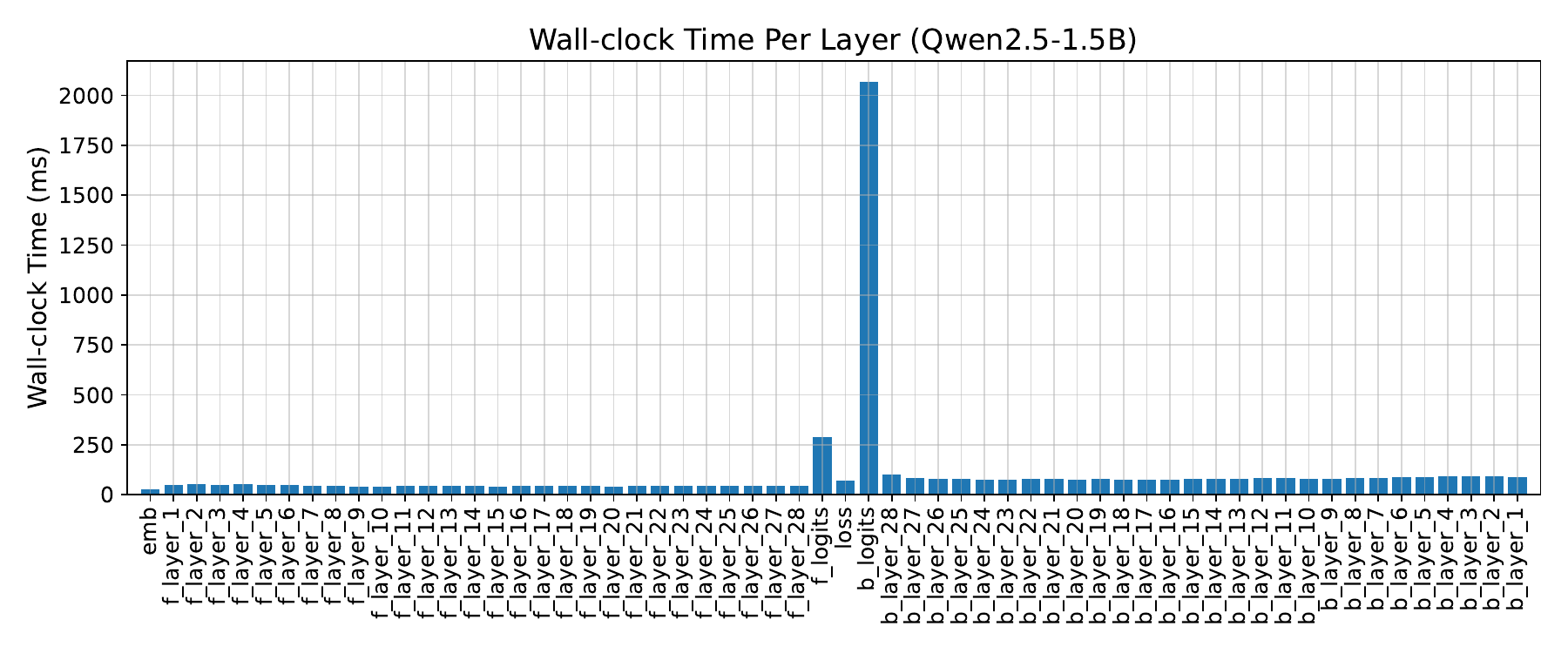}
    \caption{Per-layer memory footprint and wall-clock time. On the x-axis, \texttt{emb} stands for the embedding layer; layer name starts with \texttt{f} stands for forward and \texttt{b} for backward.}
    \label{fig:per_layer}
\end{figure*}

\paragraph{Decompression overhead.}
Table~\ref{tab:decomp} shows the decompression overhead for the forward and backward passes across different LLMs.
Decompression accounts for 32\% to 42\% of the time in the forward pass, and 13\% to 24\% in the backward pass, as the backward pass involves additional operations for gradient computation.
Although the overall compute time increases due to decompression in each pass, the memory savings are more significant as there is no need to store the uncompressed base model weights in memory, which range from 1 to 8GB for the LLMs evaluated.

\paragraph{Impact of sequence lengths.}
Sequence length can also impact performance metrics.
We experiment with Qwen2.5 1.5B using sequence lengths of 128, 256, 512, and 1024, and summarize the results in Table~\ref{tab:seq}.
As sequence length increases, both compute time and memory footprint also increase due to the heavier computation workload.
This suggests that data sources with shorter sequences, such as messages, brief emails, and user instruction prompts, are more suitable for fine-tuning on mobile devices.
We leave the investigation of efficient fine-tuning on longer sequences on mobile devices to future work.

\paragraph{Per layer performance.}
Figure~\ref{fig:per_layer} reports the per-layer(-checkpoint) performance metrics on Qwen2.5-1.5B.
For the transformer layers, the backward pass uses approximately 50\% more memory and is 30\% slower than the corresponding forward pass.
The memory bottleneck occurs at the final linear layer and the loss layer, consistent with observations in previous work~\cite{cce}.
The compute time bottleneck is also at the final linear layer, where computing the logits and their gradients involves matrix multiplication between two very large matrices (the embeddings and the sequence logits).
Both the compute time and memory footprint of the loss function and final linear layer can potentially be optimized using fused kernels~\cite{cce} or techniques such as sampled softmax~\citep{sampled_softmax}.
Another promising direction is hardware-specific implementation, such as the 1.58-bit LLM~\citep{ma2024era}, which replaces floating-point addition and multiplication with integer addition.
We leave the exploration of these techniques to future work.

For fine-tuning non-generative tasks, where the final layer does not involve heavy matrix multiplication, both compute time and memory footprint can be further reduced, shifting the bottleneck to the transformer layers instead.

\section{Conclusion}
We propose MeBP, a memory-efficient backpropagation method for fine-tuning LoRA adapters of LLMs on-device. 
Built on gradient checkpointing, MeBP incorporates memory optimizations such as lazy weight decompression and memory-mapped activations to enable exact gradient computation with better memory–compute trade-offs. 
Compared to ZO methods, MeBP achieves significantly faster convergence and better model utility, while maintaining a memory footprint comparable to MeZO on mobile devices. 
We validate MeBP on LLMs suitable for on-device deployment, demonstrating the feasibility of practical first-order fine-tuning of LLMs under tight memory constraints.

\section*{Limitations}
Due to limited device availability, MeBP has only been verified on iOS using an iPhone 15 Pro Max. It requires the capabilities of the A17 Pro chip or newer. Performance metrics may vary on other mobile operating systems or hardware configurations.

For language modeling tasks, MeBP encounters a bottleneck at the final layer due to a large matrix multiplication, resulting in increased training time. Additionally, the current implementation does not scale well with sequence length, limiting its applicability to data types that inherently involve shorter inputs.

\bibliography{custom}

\begin{thebibliography}{29}
\providecommand{\natexlab}[1]{#1}

\bibitem[{Chen et~al.(2016)Chen, Xu, Zhang, and Guestrin}]{chen2016training}
Tianqi Chen, Bing Xu, Chiyuan Zhang, and Carlos Guestrin. 2016.
\newblock Training deep nets with sublinear memory cost.
\newblock \emph{arXiv preprint arXiv:1604.06174}.

\bibitem[{Dang et~al.(2025)Dang, Guo, Zhao, Ye, Zheng, Dai, and Tsang}]{dang2025fzoo}
Sizhe Dang, Yangyang Guo, Yanjun Zhao, Haishan Ye, Xiaodong Zheng, Guang Dai, and Ivor Tsang. 2025.
\newblock Fzoo: Fast zeroth-order optimizer for fine-tuning large language models towards adam-scale speed.
\newblock \emph{arXiv preprint arXiv:2506.09034}.

\bibitem[{Dettmers et~al.(2022)Dettmers, Lewis, Shleifer, and Zettlemoyer}]{dettmers2022bit}
Tim Dettmers, Mike Lewis, Sam Shleifer, and Luke Zettlemoyer. 2022.
\newblock 8-bit optimizers via block-wise quantization.
\newblock In \emph{International Conference on Learning Representations}.

\bibitem[{Dettmers et~al.(2023)Dettmers, Pagnoni, Holtzman, and Zettlemoyer}]{dettmers2023qlora}
Tim Dettmers, Artidoro Pagnoni, Ari Holtzman, and Luke Zettlemoyer. 2023.
\newblock {QL}o{RA}: Efficient finetuning of quantized {LLM}s.
\newblock In \emph{Thirty-seventh Conference on Neural Information Processing Systems}.

\bibitem[{developer.apple.com()}]{apple}
developer.apple.com.
\newblock Identifying high-memory use with jetsam event reports.
\newblock \url{https://developer.apple.com/documentation/xcode/identifying-high-memory-use-with-jetsam-event-reports}.
\newblock Accessed: 2025-07-01.

\bibitem[{Gemini-Team et~al.(2025)Gemini-Team, Anil, Borgeaud, Alayrac, Yu, Soricut, Schalkwyk, Dai, Hauth, Millican, Silver, Johnson, Antonoglou, Schrittwieser, Glaese, Chen, Pitler, Lillicrap, Lazaridou, Firat, Molloy, Isard, Barham, Hennigan, Lee, Viola, Reynolds, Xu, Doherty, Collins, Meyer, Rutherford, Moreira, Ayoub, Goel, Krawczyk, Du, Chi, Cheng, Ni, Shah, Kane, Chan, Faruqui, Severyn, Lin, Li, Cheng, Ittycheriah, Mahdieh, Chen, Sun, Tran, Bagri, Lakshminarayanan, Liu, Orban, Güra, Zhou, Song, Boffy, Ganapathy, Zheng, Choe, Ágoston Weisz, Zhu, Lu, Gopal, Kahn, Kula, Pitman, Shah, Taropa, Merey, Baeuml, Chen, Shafey, Zhang, Sercinoglu, Tucker, Piqueras, Krikun, Barr, Savinov, Danihelka, Roelofs, White, Andreassen, von Glehn, Yagati, Kazemi, Gonzalez, Khalman, Sygnowski, Frechette, Smith, Culp, Proleev, Luan, Chen, Lottes, Schucher, Lebron, Rrustemi, Clay, Crone, Kocisky, Zhao, Perz, Yu, Howard, Bloniarz, Rae, Lu, Sifre, Maggioni, Alcober, Garrette, Barnes, Thakoor, Austin, Barth-Maron, Wong, Joshi,
  Chaabouni, Fatiha, Ahuja, Tomar, Senter, Chadwick, Kornakov, Attaluri, Iturrate, Liu, Li, Cogan, Chen, Jia, Gu, Zhang, Grimstad, Hartman, Garcia, Pillai, Devlin, Laskin, de~Las~Casas, Valter, Tao, Blanco, Badia, Reitter, Chen, Brennan, Rivera, Brin, Iqbal, Surita, Labanowski, Rao, Winkler, Parisotto, Gu, Olszewska, Addanki, Miech, Louis, Teplyashin, Brown, Catt, Balaguer, Xiang, Wang, Ashwood, Briukhov, Webson, Ganapathy, Sanghavi, Kannan, Chang, Stjerngren, Djolonga, Sun, Bapna, Aitchison, Pejman, Michalewski, Yu, Wang, Love, Ahn, Bloxwich, Han, Humphreys, Sellam, Bradbury, Godbole, Samangooei, Damoc, Kaskasoli, Arnold, Vasudevan, Agrawal, Riesa, Lepikhin, Tanburn, Srinivasan, Lim, Hodkinson, Shyam, Ferret, Hand, Garg, Paine, Li, Li, Giang, Neitz, Abbas, York, Reid, Cole, Chowdhery, Das, Rogozińska, Nikolaev, Sprechmann, Nado, Zilka, Prost, He, Monteiro, Mishra, Welty, Newlan, Jia, Allamanis, Hu, de~Liedekerke, Gilmer, Saroufim, Rijhwani, Hou, Shrivastava, Baddepudi, Goldin, Ozturel, Cassirer, Xu, Sohn,
  Sachan, Amplayo, Swanson, Petrova, Narayan, Guez, Brahma, Landon, Patel, Zhao, Villela, Wang, Jia, Rahtz, Giménez, Yeung, Keeling, Georgiev, Mincu, Wu, Haykal, Saputro, Vodrahalli, Qin, Cankara, Sharma, Fernando, Hawkins, Neyshabur, Kim, Hutter, Agrawal, Castro-Ros, van~den Driessche, Wang, Yang, yiin Chang, Komarek, McIlroy, Lučić, Zhang, Farhan, Sharman, Natsev, Michel, Bansal, Qiao, Cao, Shakeri, Butterfield, Chung, Rubenstein, Agrawal, Mensch, Soparkar, Lenc, Chung, Pope, Maggiore, Kay, Jhakra, Wang, Maynez, Phuong, Tobin, Tacchetti, Trebacz, Robinson, Katariya, Riedel, Bailey, Xiao, Ghelani, Aroyo, Slone, Houlsby, Xiong, Yang, Gribovskaya, Adler, Wirth, Lee, Li, Kagohara, Pavagadhi, Bridgers, Bortsova, Ghemawat, Ahmed, Liu, Powell, Bolina, Iinuma, Zablotskaia, Besley, Chung, Dozat, Comanescu, Si, Greer, Su, Polacek, Kaufman, Tokumine, Hu, Buchatskaya, Miao, Elhawaty, Siddhant, Tomasev, Xing, Greer, Miller, Ashraf, Roy, Zhang, Ma, Filos, Besta, Blevins, Klimenko, Yeh, Changpinyo, Mu, Chang,
  Pajarskas, Muir, Cohen, Lan, Haridasan, Marathe, Hansen, Douglas, Samuel, Wang, Austin, Lan, Jiang, Chiu, Lorenzo, Sjösund, Cevey, Gleicher, Avrahami, Boral, Srinivasan, Selo, May, Aisopos, Hussenot, Soares, Baumli, Chang, Recasens, Caine, Pritzel, Pavetic, Pardo, Gergely, Frye, Ramasesh, Horgan, Badola, Kassner, Roy, Dyer, Campos, Tomala, Tang, Badawy, White, Mustafa, Lang, Jindal, Vikram, Gong, Caelles, Hemsley, Thornton, Feng, Stokowiec, Zheng, Thacker, Çağlar Ünlü, Zhang, Saleh, Svensson, Bileschi, Patil, Anand, Ring, Tsihlas, Vezer, Selvi, Shevlane, Rodriguez, Kwiatkowski, Daruki, Rong, Dafoe, FitzGerald, Gu-Lemberg, Khan, Hendricks, Pellat, Feinberg, Cobon-Kerr, Sainath, Rauh, Hashemi, Ives, Hasson, Noland, Cao, Byrd, Hou, Wang, Sottiaux, Paganini, Lespiau, Moufarek, Hassan, Shivakumar, van Amersfoort, Mandhane, Joshi, Goyal, Tung, Brock, Sheahan, Misra, Li, Rakićević, Dehghani, Liu, Mittal, Oh, Noury, Sezener, Huot, Lamm, Cao, Chen, Mudgal, Stella, Brooks, Vasudevan, Liu, Chain, Melinkeri,
  Cohen, Wang, Seymore, Zubkov, Goel, Yue, Krishnakumaran, Albert, Hurley, Sano, Mohananey, Joughin, Filonov, Kępa, Eldawy, Lim, Rishi, Badiezadegan, Bos, Chang, Jain, Padmanabhan, Puttagunta, Krishna, Baker, Kalb, Bedapudi, Kurzrok, Lei, Yu, Litvin, Zhou, Wu, Sobell, Siciliano, Papir, Neale, Bragagnolo, Toor, Chen, Anklin, Wang, Feng, Gholami, Ling, Liu, Walter, Moghaddam, Kishore, Adamek, Mercado, Mallinson, Wandekar, Cagle, Ofek, Garrido, Lombriser, Mukha, Sun, Mohammad, Matak, Qian, Peswani, Janus, Yuan, Schelin, David, Garg, He, Duzhyi, Älgmyr, Lottaz, Li, Yadav, Xu, Chinien, Shivanna, Chuklin, Li, Spadine, Wolfe, Mohamed, Das, Dai, He, von Dincklage, Upadhyay, Maurya, Chi, Krause, Salama, Rabinovitch, M, Selvan, Dektiarev, Ghiasi, Guven, Gupta, Liu, Sharma, Shtacher, Paul, Akerlund, Aubet, Huang, Zhu, Zhu, Teixeira, Fritze, Bertolini, Marinescu, Bölle, Paulus, Gupta, Latkar, Chang, Sanders, Wilson, Wu, Tan, Thiet, Doshi, Lall, Mishra, Chen, Luong, Benjamin, Lee, Andrejczuk, Rabiej, Ranjan, Styrc,
  Yin, Simon, Harriott, Bansal, Robsky, Bacon, Greene, Mirylenka, Zhou, Sarvana, Goyal, Andermatt, Siegler, Horn, Israel, Pongetti, Chen, Selvatici, Silva, Wang, Tolins, Guu, Yogev, Cai, Agostini, Shah, Nguyen, Donnaile, Pereira, Friso, Stambler, Kurzrok, Kuang, Romanikhin, Geller, Yan, Jang, Lee, Fica, Malmi, Tan, Banica, Balle, Pham, Huang, Avram, Shi, Singh, Hidey, Ahuja, Saxena, Dooley, Potharaju, O'Neill, Gokulchandran, Foley, Zhao, Dusenberry, Liu, Mehta, Kotikalapudi, Safranek-Shrader, Goodman, Kessinger, Globen, Kolhar, Gorgolewski, Ibrahim, Song, Eichenbaum, Brovelli, Potluri, Lahoti, Baetu, Ghorbani, Chen, Crawford, Pal, Sridhar, Gurita, Mujika, Petrovski, Cedoz, Li, Chen, Santo, Goyal, Punjabi, Kappaganthu, Kwak, LV, Velury, Choudhury, Hall, Shah, Figueira, Thomas, Lu, Zhou, Kumar, Jurdi, Chikkerur, Ma, Yu, Kwak, Ähdel, Rajayogam, Choma, Liu, Barua, Ji, Park, Hellendoorn, Bailey, Bilal, Zhou, Khatir, Sutton, Rzadkowski, Macintosh, Vij, Shagin, Medina, Liang, Zhou, Shah, Bi, Dankovics, Banga,
  Lehmann, Bredesen, Lin, Hoffmann, Lai, Chung, Yang, Balani, Bražinskas, Sozanschi, Hayes, Alcalde, Makarov, Chen, Stella, Snijders, Mandl, Kärrman, Nowak, Wu, Dyck, Vaidyanathan, R, Mallet, Rudominer, Johnston, Mittal, Udathu, Christensen, Verma, Irving, Santucci, Elsayed, Davoodi, Georgiev, Tenney, Hua, Cideron, Leurent, Alnahlawi, Georgescu, Wei, Zheng, Scandinaro, Jiang, Snoek, Sundararajan, Wang, Ontiveros, Karo, Cole, Rajashekhar, Tumeh, Ben-David, Jain, Uesato, Datta, Bunyan, Wu, Zhang, Stanczyk, Zhang, Steiner, Naskar, Azzam, Johnson, Paszke, Chiu, Elias, Mohiuddin, Muhammad, Miao, Lee, Vieillard, Park, Zhang, Stanway, Garmon, Karmarkar, Dong, Lee, Kumar, Zhou, Evens, Isaac, Irving, Loper, Fink, Arkatkar, Chen, Shafran, Petrychenko, Chen, Jia, Levskaya, Zhu, Grabowski, Mao, Magni, Yao, Snaider, Casagrande, Palmer, Suganthan, Castaño, Giannoumis, Kim, Rybiński, Sreevatsa, Prendki, Soergel, Goedeckemeyer, Gierke, Jafari, Gaba, Wiesner, Wright, Wei, Vashisht, Kulizhskaya, Hoover, Le, Li, Iwuanyanwu,
  Liu, Ramirez, Khorlin, Cui, LIN, Wu, Aguilar, Pallo, Chakladar, Perng, Abellan, Zhang, Dasgupta, Kushman, Penchev, Repina, Wu, van~der Weide, Ponnapalli, Kaplan, Simsa, Li, Dousse, Yang, Piper, Ie, Pasumarthi, Lintz, Vijayakumar, Andor, Valenzuela, Lui, Paduraru, Peng, Lee, Zhang, Greene, Nguyen, Kurylowicz, Hardin, Dixon, Janzer, Choo, Feng, Zhang, Singhal, Du, McKinnon, Antropova, Bolukbasi, Keller, Reid, Finchelstein, Raad, Crocker, Hawkins, Dadashi, Gaffney, Franko, Bulanova, Leblond, Chung, Askham, Cobo, Xu, Fischer, Xu, Sorokin, Alberti, Lin, Evans, Dimitriev, Forbes, Banarse, Tung, Omernick, Bishop, Sterneck, Jain, Xia, Amid, Piccinno, Wang, Banzal, Mankowitz, Polozov, Krakovna, Brown, Bateni, Duan, Firoiu, Thotakuri, Natan, Geist, tan Girgin, Li, Ye, Roval, Tojo, Kwong, Lee-Thorp, Yew, Sinopalnikov, Ramos, Mellor, Sharma, Wu, Miller, Sonnerat, Vnukov, Greig, Beattie, Caveness, Bai, Eisenschlos, Korchemniy, Tsai, Jasarevic, Kong, Dao, Zheng, Liu, Yang, Zhu, Teh, Sanmiya, Gladchenko, Trdin, Toyama,
  Rosen, Tavakkol, Xue, Elkind, Woodman, Carpenter, Papamakarios, Kemp, Kafle, Grunina, Sinha, Talbert, Wu, Owusu-Afriyie, Du, Thornton, Pont-Tuset, Narayana, Li, Fatehi, Wieting, Ajmeri, Uria, Ko, Knight, Héliou, Niu, Gu, Pang, Li, Levine, Stolovich, Santamaria-Fernandez, Goenka, Yustalim, Strudel, Elqursh, Deck, Lee, Li, Levin, Hoffmann, Holtmann-Rice, Bachem, Arora, Koh, Yeganeh, Põder, Tariq, Sun, Ionita, Seyedhosseini, Tafti, Liu, Gulati, Liu, Ye, Chrzaszcz, Wang, Sethi, Li, Brown, Singh, Fan, Parisi, Stanton, Koverkathu, Choquette-Choo, Li, Lu, Ittycheriah, Shroff, Varadarajan, Bahargam, Willoughby, Gaddy, Desjardins, Cornero, Robenek, Mittal, Albrecht, Shenoy, Moiseev, Jacobsson, Ghaffarkhah, Rivière, Walton, Crepy, Parrish, Zhou, Farabet, Radebaugh, Srinivasan, van~der Salm, Fidjeland, Scellato, Latorre-Chimoto, Klimczak-Plucińska, Bridson, de~Cesare, Hudson, Mendolicchio, Walker, Morris, Mauger, Guseynov, Reid, Odoom, Loher, Cotruta, Yenugula, Grewe, Petrushkina, Duerig, Sanchez, Yadlowsky, Shen,
  Globerson, Webb, Dua, Li, Bhupatiraju, Hurt, Qureshi, Agarwal, Shani, Eyal, Khare, Belle, Wang, Tekur, Kale, Wei, Sang, Saeta, Liechty, Sun, Zhao, Lee, Nayak, Fritz, Vuyyuru, Aslanides, Vyas, Wicke, Ma, Eltyshev, Martin, Cate, Manyika, Amiri, Kim, Xiong, Kang, Luisier, Tripuraneni, Madras, Guo, Waters, Wang, Ainslie, Baldridge, Zhang, Pruthi, Bauer, Yang, Mansour, Gelman, Xu, Polovets, Liu, Cai, Chen, Sheng, Xue, Ozair, Angermueller, Li, Sinha, Wang, Wiesinger, Koukoumidis, Tian, Iyer, Gurumurthy, Goldenson, Shah, Blake, Yu, Urbanowicz, Palomaki, Fernando, Durden, Mehta, Momchev, Rahimtoroghi, Georgaki, Raul, Ruder, Redshaw, Lee, Zhou, Jalan, Li, Hechtman, Schuh, Nasr, Milan, Mikulik, Franco, Green, Nguyen, Kelley, Mahendru, Hu, Howland, Vargas, Hui, Bansal, Rao, Ghiya, Wang, Ye, Sarr, Preston, Elish, Li, Kaku, Gupta, Pasupat, Juan, Someswar, M., Chen, Amini, Fabrikant, Chu, Dong, Muthal, Buthpitiya, Jauhari, Hua, Khandelwal, Hitron, Ren, Rinaldi, Drath, Dabush, Jiang, Godhia, Sachs, Chen, Fan, Taitelbaum,
  Noga, Dai, Wang, Liang, Hamer, Ferng, Elkind, Atias, Lee, Listík, Carlen, van~de Kerkhof, Pikus, Zaher, Müller, Zykova, Stefanec, Gatsko, Hirnschall, Sethi, Xu, Ahuja, Tsai, Stefanoiu, Feng, Dhandhania, Katyal, Gupta, Parulekar, Pitta, Zhao, Bhatia, Bhavnani, Alhadlaq, Li, Danenberg, Tu, Pine, Filippova, Ghosh, Limonchik, Urala, Lanka, Clive, Sun, Li, Wu, Hongtongsak, Li, Thakkar, Omarov, Majmundar, Alverson, Kucharski, Patel, Jain, Zabelin, Pelagatti, Kohli, Kumar, Kim, Sankar, Shah, Ramachandruni, Zeng, Bariach, Weidinger, Vu, Andreev, He, Hui, Kashem, Subramanya, Hsiao, Hassabis, Kavukcuoglu, Sadovsky, Le, Strohman, Wu, Petrov, Dean, and Vinyals}]{geminiteam2025geminifamilyhighlycapable}
Gemini-Team, Rohan Anil, Sebastian Borgeaud, Jean-Baptiste Alayrac, Jiahui Yu, Radu Soricut, Johan Schalkwyk, Andrew~M. Dai, Anja Hauth, Katie Millican, David Silver, Melvin Johnson, Ioannis Antonoglou, Julian Schrittwieser, Amelia Glaese, Jilin Chen, Emily Pitler, Timothy Lillicrap, Angeliki Lazaridou, and 1332 others. 2025.
\newblock Gemini: A family of highly capable multimodal models.

\bibitem[{Gemma-Team et~al.(2025)Gemma-Team, Kamath, Ferret, Pathak, Vieillard, Merhej, Perrin, Matejovicova, Ramé, Rivière, Rouillard, Mesnard, Cideron, bastien Grill, Ramos, Yvinec, Casbon, Pot, Penchev, Liu, Visin, Kenealy, Beyer, Zhai, Tsitsulin, Busa-Fekete, Feng, Sachdeva, Coleman, Gao, Mustafa, Barr, Parisotto, Tian, Eyal, Cherry, Peter, Sinopalnikov, Bhupatiraju, Agarwal, Kazemi, Malkin, Kumar, Vilar, Brusilovsky, Luo, Steiner, Friesen, Sharma, Sharma, Gilady, Goedeckemeyer, Saade, Feng, Kolesnikov, Bendebury, Abdagic, Vadi, György, Pinto, Das, Bapna, Miech, Yang, Paterson, Shenoy, Chakrabarti, Piot, Wu, Shahriari, Petrini, Chen, Lan, Choquette-Choo, Carey, Brick, Deutsch, Eisenbud, Cattle, Cheng, Paparas, Sreepathihalli, Reid, Tran, Zelle, Noland, Huizenga, Kharitonov, Liu, Amirkhanyan, Cameron, Hashemi, Klimczak-Plucińska, Singh, Mehta, Lehri, Hazimeh, Ballantyne, Szpektor, Nardini, Pouget-Abadie, Chan, Stanton, Wieting, Lai, Orbay, Fernandez, Newlan, yeong Ji, Singh, Black, Yu, Hui, Vodrahalli,
  Greff, Qiu, Valentine, Coelho, Ritter, Hoffman, Watson, Chaturvedi, Moynihan, Ma, Babar, Noy, Byrd, Roy, Momchev, Chauhan, Sachdeva, Bunyan, Botarda, Caron, Rubenstein, Culliton, Schmid, Sessa, Xu, Stanczyk, Tafti, Shivanna, Wu, Pan, Rokni, Willoughby, Vallu, Mullins, Jerome, Smoot, Girgin, Iqbal, Reddy, Sheth, Põder, Bhatnagar, Panyam, Eiger, Zhang, Liu, Yacovone, Liechty, Kalra, Evci, Misra, Roseberry, Feinberg, Kolesnikov, Han, Kwon, Chen, Chow, Zhu, Wei, Egyed, Cotruta, Giang, Kirk, Rao, Black, Babar, Lo, Moreira, Martins, Sanseviero, Gonzalez, Gleicher, Warkentin, Mirrokni, Senter, Collins, Barral, Ghahramani, Hadsell, Matias, Sculley, Petrov, Fiedel, Shazeer, Vinyals, Dean, Hassabis, Kavukcuoglu, Farabet, Buchatskaya, Alayrac, Anil, Dmitry, Lepikhin, Borgeaud, Bachem, Joulin, Andreev, Hardin, Dadashi, and Hussenot}]{team2025gemma}
Gemma-Team, Aishwarya Kamath, Johan Ferret, Shreya Pathak, Nino Vieillard, Ramona Merhej, Sarah Perrin, Tatiana Matejovicova, Alexandre Ramé, Morgane Rivière, Louis Rouillard, Thomas Mesnard, Geoffrey Cideron, Jean bastien Grill, Sabela Ramos, Edouard Yvinec, Michelle Casbon, Etienne Pot, Ivo Penchev, and 197 others. 2025.
\newblock Gemma 3 technical report.
\newblock \emph{arXiv preprint arXiv:2503.19786}.

\bibitem[{Gunter et~al.(2024)Gunter, Wang, Wang, Pang, Narayanan, Zhang, Zhang, Chen, Chiu, Qiu et~al.}]{gunter2024apple}
Tom Gunter, Zirui Wang, Chong Wang, Ruoming Pang, Andy Narayanan, Aonan Zhang, Bowen Zhang, Chen Chen, Chung-Cheng Chiu, David Qiu, and 1 others. 2024.
\newblock Apple intelligence foundation language models.
\newblock \emph{arXiv preprint arXiv:2407.21075}.

\bibitem[{Hu et~al.(2022)Hu, yelong shen, Wallis, Allen-Zhu, Li, Wang, Wang, and Chen}]{hu2022lora}
Edward~J Hu, yelong shen, Phillip Wallis, Zeyuan Allen-Zhu, Yuanzhi Li, Shean Wang, Lu~Wang, and Weizhu Chen. 2022.
\newblock Lo{RA}: Low-rank adaptation of large language models.
\newblock In \emph{International Conference on Learning Representations}.

\bibitem[{Jean et~al.(2015)Jean, Cho, Memisevic, and Bengio}]{sampled_softmax}
S{\'e}bastien Jean, Kyunghyun Cho, Roland Memisevic, and Yoshua Bengio. 2015.
\newblock \href {https://doi.org/10.3115/v1/P15-1001} {On using very large target vocabulary for neural machine translation}.
\newblock In \emph{Proceedings of the 53rd Annual Meeting of the Association for Computational Linguistics and the 7th International Joint Conference on Natural Language Processing (Volume 1: Long Papers)}, pages 1--10.

\bibitem[{Kairouz et~al.(2021)Kairouz, McMahan, Avent, Bellet, Bennis, Bhagoji, Bonawitz, Charles, Cormode, Cummings et~al.}]{kairouz2021advances}
Peter Kairouz, H~Brendan McMahan, Brendan Avent, Aur{\'e}lien Bellet, Mehdi Bennis, Arjun~Nitin Bhagoji, Kallista Bonawitz, Zachary Charles, Graham Cormode, Rachel Cummings, and 1 others. 2021.
\newblock Advances and open problems in federated learning.
\newblock \emph{Foundations and trends{\textregistered} in machine learning}, 14(1--2):1--210.

\bibitem[{Kingma and Ba(2014)}]{kingma2014adam}
Diederik~P Kingma and Jimmy Ba. 2014.
\newblock Adam: A method for stochastic optimization.
\newblock \emph{arXiv preprint arXiv:1412.6980}.

\bibitem[{Lin et~al.(2022)Lin, Zhu, Chen, Wang, Gan, and Han}]{lin2022ondevice}
Ji~Lin, Ligeng Zhu, Wei-Ming Chen, Wei-Chen Wang, Chuang Gan, and Song Han. 2022.
\newblock On-device training under 256kb memory.
\newblock In \emph{Annual Conference on Neural Information Processing Systems (NeurIPS)}.

\bibitem[{Liu et~al.(2025)Liu, Zhao, Huang, Chen, Zhang, Zhao, Roy, Jin, Xiong, Shi et~al.}]{liu2025paretoq}
Zechun Liu, Changsheng Zhao, Hanxian Huang, Sijia Chen, Jing Zhang, Jiawei Zhao, Scott Roy, Lisa Jin, Yunyang Xiong, Yangyang Shi, and 1 others. 2025.
\newblock Paretoq: Scaling laws in extremely low-bit llm quantization.
\newblock \emph{arXiv preprint arXiv:2502.02631}.

\bibitem[{Lv et~al.(2024)Lv, Yang, Liu, Guo, and Qiu}]{lv2024lomo}
Kai Lv, Yuqing Yang, Tengxiao Liu, Qipeng Guo, and Xipeng Qiu. 2024.
\newblock \href {https://doi.org/10.18653/v1/2024.acl-long.445} {Full parameter fine-tuning for large language models with limited resources}.
\newblock In \emph{Proceedings of the 62nd Annual Meeting of the Association for Computational Linguistics (Volume 1: Long Papers)}, pages 8187--8198. Association for Computational Linguistics.

\bibitem[{Ma et~al.(2024)Ma, Wang, Ma, Wang, Wang, Huang, Dong, Wang, Xue, and Wei}]{ma2024era}
Shuming Ma, Hongyu Wang, Lingxiao Ma, Lei Wang, Wenhui Wang, Shaohan Huang, Li~Dong, Ruiping Wang, Jilong Xue, and Furu Wei. 2024.
\newblock The era of 1-bit llms: All large language models are in 1.58 bits.
\newblock \emph{arXiv preprint arXiv:2402.17764}.

\bibitem[{Malladi et~al.(2023)Malladi, Gao, Nichani, Damian, Lee, Chen, and Arora}]{malladi2023finetuning}
Sadhika Malladi, Tianyu Gao, Eshaan Nichani, Alex Damian, Jason~D. Lee, Danqi Chen, and Sanjeev Arora. 2023.
\newblock Fine-tuning language models with just forward passes.
\newblock In \emph{Advances in Neural Information Processing Systems}.

\bibitem[{McMahan et~al.(2017)McMahan, Moore, Ramage, Hampson, and y~Arcas}]{mcmahan2017communication}
Brendan McMahan, Eider Moore, Daniel Ramage, Seth Hampson, and Blaise~Aguera y~Arcas. 2017.
\newblock Communication-efficient learning of deep networks from decentralized data.
\newblock In \emph{Artificial intelligence and statistics}, pages 1273--1282. PMLR.

\bibitem[{Merity et~al.(2017)Merity, Xiong, Bradbury, and Socher}]{merity2017pointer}
Stephen Merity, Caiming Xiong, James Bradbury, and Richard Socher. 2017.
\newblock \href {https://openreview.net/forum?id=Byj72udxe} {Pointer sentinel mixture models}.
\newblock In \emph{International Conference on Learning Representations}.

\bibitem[{Peng et~al.(2024)Peng, Fu, and Wang}]{peng2024pocketllm}
Dan Peng, Zhihui Fu, and Jun Wang. 2024.
\newblock Pocketllm: Enabling on-device fine-tuning for personalized llms.
\newblock \emph{arXiv preprint arXiv:2407.01031}.

\bibitem[{Qin et~al.(2024)Qin, Chen, Qian, Ding, Li, and Deng}]{qin2024federated}
Zhen Qin, Daoyuan Chen, Bingchen Qian, Bolin Ding, Yaliang Li, and Shuiguang Deng. 2024.
\newblock \href {https://openreview.net/forum?id=cit0hg4sEz} {Federated full-parameter tuning of billion-sized language models with communication cost under 18 kilobytes}.
\newblock In \emph{Forty-first International Conference on Machine Learning}.

\bibitem[{Qwen-Team et~al.(2025)Qwen-Team, Yang, Yang, Zhang, Hui, Zheng, Yu, Li, Liu, Huang, Wei, Lin, Yang, Tu, Zhang, Yang, Yang, Zhou, Lin, Dang, Lu, Bao, Yang, Yu, Li, Xue, Zhang, Zhu, Men, Lin, Li, Tang, Xia, Ren, Ren, Fan, Su, Zhang, Wan, Liu, Cui, Zhang, and Qiu}]{qwen2024qwen25technicalreport}
Qwen-Team, An~Yang, Baosong Yang, Beichen Zhang, Binyuan Hui, Bo~Zheng, Bowen Yu, Chengyuan Li, Dayiheng Liu, Fei Huang, Haoran Wei, Huan Lin, Jian Yang, Jianhong Tu, Jianwei Zhang, Jianxin Yang, Jiaxi Yang, Jingren Zhou, Junyang Lin, and 24 others. 2025.
\newblock Qwen2.5 technical report.
\newblock \emph{arXiv preprint arXiv:2412.15115}.

\bibitem[{Rajbhandari et~al.(2020)Rajbhandari, Rasley, Ruwase, and He}]{zero}
Samyam Rajbhandari, Jeff Rasley, Olatunji Ruwase, and Yuxiong He. 2020.
\newblock Zero: memory optimizations toward training trillion parameter models.
\newblock In \emph{Proceedings of the International Conference for High Performance Computing, Networking, Storage and Analysis}, SC '20. IEEE Press.

\bibitem[{source.android.com()}]{android}
source.android.com.
\newblock Low memory killer daemon.
\newblock \url{https://source.android.com/docs/core/perf/lmkd}.
\newblock Accessed: 2025-07-01.

\bibitem[{Spall(1992)}]{Spall1992MultivariateSA}
James~C. Spall. 1992.
\newblock Multivariate stochastic approximation using a simultaneous perturbation gradient approximation.
\newblock \emph{IEEE Transactions on Automatic Control}, 37:332--341.

\bibitem[{Wijmans et~al.(2025)Wijmans, Huval, Hertzberg, Koltun, and Kraehenbuehl}]{cce}
Erik Wijmans, Brody Huval, Alexander Hertzberg, Vladlen Koltun, and Philipp Kraehenbuehl. 2025.
\newblock \href {https://openreview.net/forum?id=E4Fk3YuG56} {Cut your losses in large-vocabulary language models}.
\newblock In \emph{The Thirteenth International Conference on Learning Representations}.

\bibitem[{Zhang et~al.(2022)Zhang, Roller, Goyal, Artetxe, Chen, Chen, Dewan, Diab, Li, Lin et~al.}]{zhang2022opt}
Susan Zhang, Stephen Roller, Naman Goyal, Mikel Artetxe, Moya Chen, Shuohui Chen, Christopher Dewan, Mona Diab, Xian Li, Xi~Victoria Lin, and 1 others. 2022.
\newblock Opt: Open pre-trained transformer language models.
\newblock \emph{arXiv preprint arXiv:2205.01068}.

\bibitem[{Zhao et~al.(2024)Zhao, Zhang, Chen, Wang, Anandkumar, and Tian}]{zhao2024galore}
Jiawei Zhao, Zhenyu Zhang, Beidi Chen, Zhangyang Wang, Anima Anandkumar, and Yuandong Tian. 2024.
\newblock Galore: Memory-efficient {LLM} training by gradient low-rank projection.
\newblock In \emph{Forty-first International Conference on Machine Learning}.

\bibitem[{Zhao et~al.(2025)Zhao, Dang, Ye, Dai, Qian, and Tsang}]{zhao2025secondorder}
Yanjun Zhao, Sizhe Dang, Haishan Ye, Guang Dai, Yi~Qian, and Ivor Tsang. 2025.
\newblock \href {https://openreview.net/forum?id=bEqI61iBue} {Second-order fine-tuning without pain for {LLM}s: A hessian informed zeroth-order optimizer}.
\newblock In \emph{The Thirteenth International Conference on Learning Representations}.

\end{thebibliography}

\appendix
\newpage
\section{Improved ZO Methods}
\label{appendix:zo}

\begin{figure}[t]
\centering
    \begin{minipage}{\linewidth}
        \includegraphics[width=\linewidth]{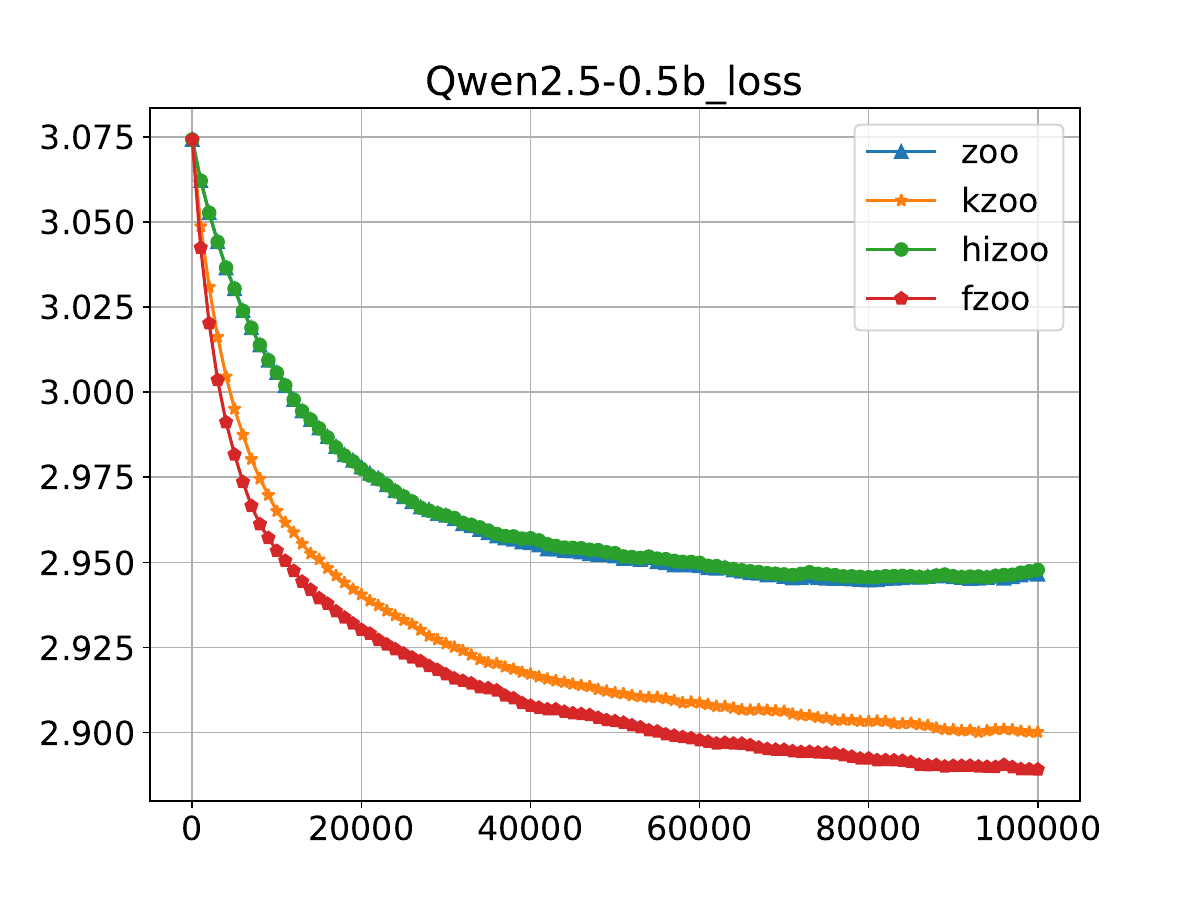}
    \end{minipage}
    \begin{minipage}{\linewidth}
        \includegraphics[width=\linewidth]{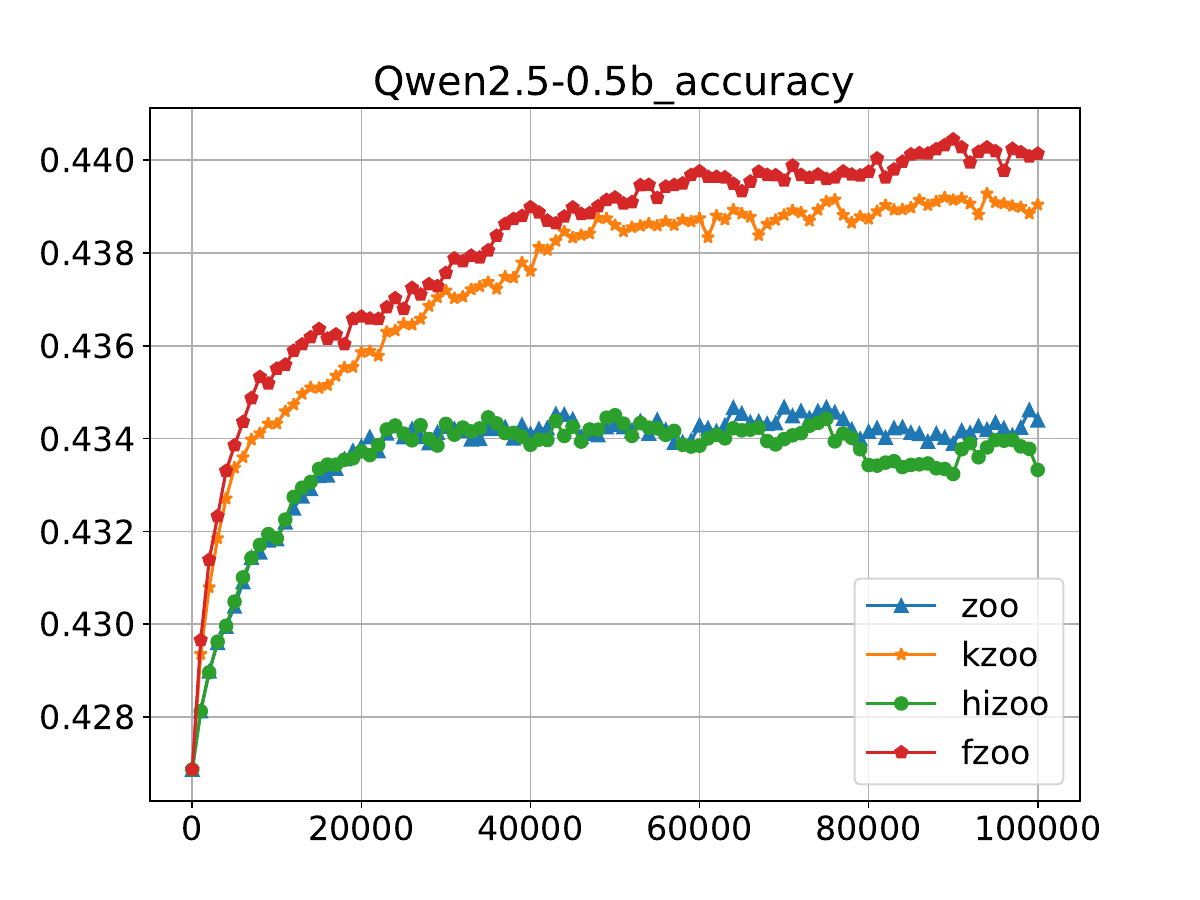}
    \end{minipage}
    \caption{The performance of improved ZO methods~(zoo~\citep{malladi2023finetuning}, kzoo~\citep{qin2024federated},hizoo~\citep{zhao2025secondorder}, fzoo~\citep{dang2025fzoo}).}
    \label{fig:zoo_improve}
\end{figure}

For improved ZO methods, \citet{qin2024federated} use more than one seed per iteration to provide better gradient estimation (KZOO).
\citet{zhao2025secondorder} leverage second-order information via the Hessian matrix (HiZOO), while \citet{dang2025fzoo} use more gradient estimations per iteration, with each estimation requiring only one forward pass rather than two (FZOO).
For fair comparison, we consider 4 gradient estimations per iteration for KZOO and 8 for FZOO.
For HiZOO, we follow the same setting as \citet{malladi2023finetuning}, using 1 gradient estimation (i.e., two forward passes).
All other experimental settings are the same as those described in Section~\ref{sec:zoo}.
We present the results in Figure~\ref{fig:zoo_improve}.
While these methods improve the convergence rate compared to vanilla ZO, they still exhibit a much slower convergence trend than the first-order (FO) method shown in Figure~\ref{fig:qwen2.5}.

\end{document}